# IMPROVING IMAGE CODING FOR MACHINES THROUGH OPTIMIZING ENCODER VIA AUXILIARY LOSS

*Kei Iino*[†], *Shunsuke Akamatsu*[†], *Hiroshi Watanabe*[†], *Shohei Enomoto*[‡], *Akira Sakamoto*[‡], *Takeharu Eda*[‡]

[†]Graduate School of Fundamental Science and Engineering, Waseda University, Tokyo, Japan
[‡]NTT Software Innovation Center, Tokyo, Japan

## ABSTRACT

Image coding for machines (ICM) aims to compress images for machine analysis using recognition models rather than human vision. Hence, in ICM, it is important for the encoder to recognize and compress the information necessary for the machine recognition task. There are two main approaches in learned ICM; optimization of the compression model based on task loss, and Region of Interest (ROI) based bit allocation. These approaches provide the encoder with the recognition capability. However, optimization with task loss becomes difficult when the recognition model is deep, and ROI-based methods often involve extra overhead during evaluation. In this study, we propose a novel training method for learned ICM models that applies auxiliary loss to the encoder to improve its recognition capability and rate-distortion performance. Our method achieves Bjøntegaard Delta rate improvements of 27.7% and 20.3% in object detection and semantic segmentation tasks, compared to the conventional training method.

***Index Terms***— Image coding for machines, ICM, learned image compression, auxiliary loss, VCM

## 1. INTRODUCTION

In recent years, the performance of deep neural networks (DNNs) has seen remarkable improvements, leading to their widespread use in various machine analysis systems such as video surveillance systems and speech recognition systems. The scenarios in which machine analysis systems are utilized are generally classified into edge computing and cloud computing. In edge computing, machine analysis systems such as DNNs are mounted on edge devices. Therefore, there are no transmission costs and delays. However, edge devices generally have limited resources compared to the cloud, which limits the performance and number of models that can be deployed. On the other hand, in cloud computing, machine analysis systems are deployed on the cloud, enabling the use of high-precision and multiple models. However, to receive inputs for the cloud systems, data compression and transmission are necessary on the edge devices. In image recognition systems, such as video surveillance systems, the videos captured on edge devices are compressed and encoded using hand-crafted codecs such as HEVC [1], VVC [2], etc., and then transmitted to the cloud. These codecs are optimized assuming the decoded data will be viewed by humans, meaning the decoded data have to be similar to the original data and provide high visual quality.

However, in machine analysis scenarios, the decoded data is often input into a recognition model and analyzed without human viewing. In such cases, rather than similarity to the original data or visual quality, it is required to minimize the data size during compression while minimizing the degradation of recognition performance. Image coding for machines (ICM) is the answer to these requirements [3-9]. This research field has attracted much attention in recent years as its demand has increased with the development of automated machine analysis. Video Coding for Machines (VCM) [10] in MPEG is engaged in standardization activities close to this research field. In addition, learned image compression [16,17], which has been developing remarkably in recent years, has also accelerated ICM research (learned ICM) [3-5,8,9,11-13].

In ICM, it is ideal to identify and extract only the information needed for the recognition task, and to compress it [6-9,11-13]. Therefore, the main approaches of learned ICM research can be categorized into the following two approaches. The first one is to optimize the compression model with the task loss of the recognition model [3-5]. By training with the task loss, the compression model can directly optimize the trade-off between the bitrate and the performance of the recognition model. The second one is a Region of Interest (ROI) based bit allocation scheme [6-9]. This approach designs a task-oriented encoder by assigning bits with priority to the foreground in the image. However, each of these approaches has its own problems. In the task loss approach, sufficient optimization may be difficult depending on the complexity of the task and the depth of the recognition model [11-13]. In particular, when the network of recognition models is deep, optimization of compression models located in shallow layers becomes difficult due to the nature of backpropagation [14,15]. In the ROI-based approach, additional overhead is incurred on the encoder side during evaluation for getting the ROI maps [6-9]. Furthermore, defining the ROIs can be challenging for tasks that involve background classification, such as semantic and panoptic segmentation, and for image captioning tasks.

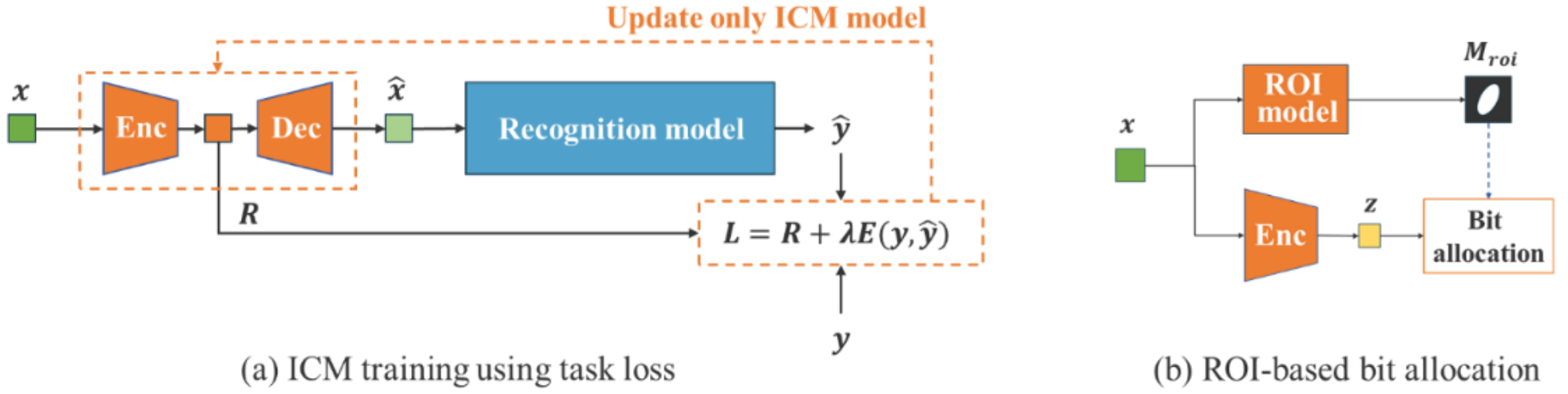

**Fig. 1** Two main approaches of learned ICM research.

In this study, we propose a novel training method for learned ICM models using auxiliary loss. Our method imposes an auxiliary loss on the encoder via a lightweight recognition model during training.

Our experiments show that our method improves the encoder's recognition capability and the ICM model's rate-distortion (R-D) performance without any additional overhead during evaluation.

## 2. RELATED WORK

### 2.1. Learned image compression

Learned image compression has attracted much attention in recent years because its compression performance has been greatly improved and even exceeds the performance of conventional hand-crafted codecs such as HEVC and VVC [16,17]. The compression model typically consists of encoder, entropy model and decoder, and the model is optimized using the following loss function [16,17]:

$$L = R + \lambda D, \tag{1}$$

where $R$ represents the bitrate estimated by the entropy model after compression, and $D$ represents the distortion calculated from the reconstructed data. Additionally, $\lambda$ is the Lagrange multiplier. Generally, image reconstruction error is used for the distortion $D$ in image coding for humans:

$$L = R + \lambda D(x, \hat{x}), \tag{2}$$

where $x$ and $\hat{x}$ denote the input and the decoded image. Mean square error (MSE) or structural similarity index measure (SSIM) is typically used as the error function $D$. By using the image reconstruction error for $D$, the similarity between the original and the decoded image is ensured. Learned ICM [3-5,8,9,11-13] is based on this learning-based image compression and is the main subject of this study.

### 2.2. Image coding for machines (ICM)

ICM aims to compress images for machine analysis using recognition models rather than human vision. The success of learned image compression as described in Sec. 2.1 has accelerated ICM research (learned ICM) [3-5,8,9,11-13]. In ICM, it is considered ideal to extract and compress information necessary for recognition tasks to minimize performance degradation and bitrate. For this reason, the main approaches of learned ICM research are generally classified into two approaches.

The first approach is optimizing the ICM model based on the task loss of the recognition model [3-5], as illustrated in Fig. 1(a). In learned ICM research the feature reconstruction errors [4] or the task loss [3-5] of the recognition model are used as D in Eq. (1). When using the feature reconstruction error, the loss function can be represented as:

$$L = R + \lambda \textstyle\sum_i D\left(f_i, \hat{f}_i\right), \tag{3}$$

where $f_i$ and $\hat{f}_i$ represent the feature maps of the i-th layer generated from the original and decoded image, respectively. MSE is typically used as $D$. By training the ICM model using this distillation manner, the features from decoded images come to resemble those from the original images, which indirectly reduces the degradation of recognition performance. A more direct approach is to use the task loss of the recognition model:

$$L = R + \lambda E(y, \hat{y}), \tag{4}$$

where $E$ represents the task loss function of the recognition model, $y$ represents the target label and $\hat{y}$ is the output of the model when the decoded data is input. This approach can directly optimize the trade-off between bitrate and the recognition model performance. Harell *et al.* [4] analyzed the R-D theory in ICM domain. According to [4], when using feature distillation for ICM training, it is desirable to compute distillation errors in the deeper layers of the recognition model. Furthermore, they showed that optimization using the task loss is optimal from the perspective of the R-D theory. Yamazaki *et al.* [5] proposed using the model output of the original image as the label $y$ in Eq. (4) instead of the ground truth (GT) label. Since GT labels for training data are rarely available in the real world, this method makes it possible to optimize ICM models with task loss in the real world. In our experiments, we followed this training manner. Note that, as in these previous studies [3-5, 8], we fixed the parameters of the recognition model and only optimized the learned ICM model.

**Fig. 2.** The overview of the proposed ICM training method using auxiliary loss.

The second approach is the ROI-based bit allocation [6-9]. In many computer vision tasks, the impact of background region on task accuracy is smaller compared to foreground region [6-9]. Therefore, several methods have been proposed to allocate more bits to the foreground region as the ROI and reduce the number of bits in the background outside the ROI as shown in Fig. 1 (b). In [6,7], the ROI-based approach is applied to traditional hand-crafted codecs, while [8,9] applied it to learned image compression models. Ahonen *et al.* [8] have proposed an ad hoc method to apply the ROI-based approach to a pretrained compression model. Specifically, the ROI mask $M_{roi}$ generated by the additional recognition model is binarized and the pixel values outside the ROI in the latent representation $z$ are divided by the quantization factor (QF) which is greater than 1. This operation suppresses the variance of the pixel values outside the ROI, thereby reducing the number of bits allocated outside the ROI.

### 2.3. Auxiliary loss

It is well known that as the depth of a network increases, optimizing the shallower layers, which means the layers near the input layer, becomes more difficult [14,15]. Auxiliary loss is one way to solve such a problem and is a loss function that is added to the main loss to help train the model [14,15]. Auxiliary loss is computed on the output of an auxiliary branch (also called auxiliary head), which is a branch from the middle layer of the model, so that errors about the target task can be propagated properly before that middle layer. As a result, the training process becomes more stable even if the model is deep, which is expected to speed up training convergence and improve the final performance. In this study, we focus on auxiliary loss for the first time in ICM training and propose its application.

## 3. PROPOSED METHOD

In ICM, it is ideal to identify and extract only the information needed for the recognition task, and to compress it. For this reason, many studies take task loss-based optimization [3-5] or ROI-based bit allocation approaches [8,9]. However, it has been reported that training using task loss can make it difficult to optimize the ICM model when the recognition model is deep [11-13]. As mentioned in Sec. 2.3, when the depth of a network increases, optimizing the shallower layers becomes more difficult. In training ICM models, such problems arise because the ICM model to be optimized is located before the input layer or shallow layer of the fixed recognition model. As a result, the encoder may not acquire sufficient recognition ability. Furthermore, many ROI-based approaches require an additional network on the encoder side to obtain the ROI mask [6-9]. However, the resources on the encoder side (usually edge devices) are generally much smaller than those on the decoder side (usually the cloud), so it is undesirable to increase the overhead on the encoder side during evaluation. In general, most ROI-based approach treats the foreground as ROI, but it is not obvious which region or what information in the image is important on the recognition task and model [18]. In particular, it is difficult to determine the ROI itself in semantic and panoptic segmentation tasks, which also classify several background classes, and in image captioning tasks, which provide explanations for images.

In this study, we propose to apply auxiliary loss when training ICM models to solve the above problems. As described in Sec. 2.3, auxiliary loss can help the training process in shallow layers. Therefore, the use of auxiliary loss is considered effective in ICM model training because the learnable ICM model is located before the input layer or shallow layer of the fixed recognition model. In the proposed method, we impose the auxiliary loss only on the encoder of the ICM model. This approach supports the encoder to acquire ROI-based-like recognition capabilities without any additional overhead during evaluation.

An overview of the proposed method is shown in Fig. 2. In the proposed method, an auxiliary branch, which is a lightweight model based on the recognition model, is set before the decoder and is jointly optimized with the ICM model. The loss function is defined as follows:

$$L = R + \lambda\{E(y, \hat{y}) + \alpha E(y, \hat{y}_{aux})\}, \quad (5)$$

where $\hat{y}_{aux}$ is the output of the auxiliary branch and $\alpha$ is the weight factor that balances the main task loss $E(y, \hat{y})$ and the auxiliary loss $E(y, \hat{y}_{aux})$. Since the auxiliary branch is placed just before the decoder, the auxiliary loss is backpropagated directly to the encoder which includes the entropy model. This approach helps the encoder understand the information necessary for the recognition task like the ROI-based method, even when the recognition model is deep. By limiting the effect of auxiliary loss to only the encoder rather than the entire ICM model, the negative impact on the main task can be reduced, this topic is discussed in Sec. 4.2. The proposed method does not require the definition of ROI, so this method can be applied to a variety of tasks. Furthermore, since the auxiliary branch is used only during training and not during evaluation, it does not incur the overhead like conventional ROI-based methods.

## 4. EXPERIMENTS

We evaluated the proposed method on the object detection and semantic segmentation tasks. As a baseline model, we used a compression model trained without auxiliary loss, and for object detection, we also used the compression model with the ROI-based method [8] applied to the baseline.

### 4.1 Main experiments

**Experimental setup:** For the object detection task, we used the pre-trained learned image compression model [17] (cheng2020attn in compressai [19]) as the ICM model and the pre-trained Fater-RCNN (ResNeXt101-FPN) [20] from the DetectronV2 [21] as the recognition model. We followed the training manner [5] and used the output of the Faster-RCNN from the original image as the target label $y$ in Eq. (5). As the dataset, we used COCO2017 [25], with 118287 images for training and 5000 for evaluation. During training, we used 256×256 cropped images. During evaluation, we resized the images to have a shorter edge of 800 and padded them to a multiple of 64 so that they could be input into the ICM model. We used Adam optimizer and trained the compression model in 80 epochs with a batch size of 16. As the initial learning rate, we set it to 0.0001, with a decay according to a polynomial decay schedule for the last 40 epochs. The overall structure of the auxiliary branch was based on Faster-RCNN, and we reduced the depth and width of the backbone model significantly. The details of each backbone model are shown in Table 1. In Eq. (5), we set $\alpha$ to 0.5 and $\lambda$ to {0.4, 0.8, 1.6, 3.2}. For the ROI-based method [8], we used the GT mask provided in the COCO dataset as the ROI mask and set QF to 1.4.

For the semantic segmentation task, we used the same compression model as in the object detection task and used DeepLabv3+ (ResNet101) [23] from the MMSegmentation [26] as the recognition model. As in the object detection task, we followed the training manner [5]. Since training with the loss function in Eq. (5) could not reduce the validation loss of the baseline model, we added the image reconstruction error to stabilize the training:

$$L = R + \lambda\{E(y, \hat{y}) + \alpha E(y, \hat{y}_{aux}) + MSE(x, \hat{x})\}. \quad (6)$$

As the dataset, we used Pascal-Context59 [27], with 4998 images for training and 5105 images for evaluation. During training and evaluation, we resized the input images to 520×520 and padded them to multiples of 64. We used Adam optimizer and trained 80 epochs with a batch size of 8. As for the initial learning rate, we set it to 0.00005 and employed polynomial decay in each epoch. We designed the auxiliary branch based on the recognition model DeepLabv3+ and changed it to a tiny backbone model like the object detection task. The details of each backbone model are shown in the table. 2. In Eq. (6), we set $\alpha$ to 0.5 and $\lambda$ to {1, 2, 4, 8}.

**Table 1.** The backbone architectures of Faster-RCNN [20] and auxiliary branch in the experiment of object detection task. The architecture of ResNext is from [22]. RB denotes the residual block proposed in [16] and ↑ denotes upsampling.

| layer name | ResNext-101(32×8d) | Auxiliary branch |
|---|---|---|
| conv1 | $7 \times 7, 64, \downarrow 2$ | RB, 128, ↑ 2<br>RB, 128<br>RB, 128, ↑ 2 |
| conv2 | $3 \times 3, MP, \downarrow 2$<br>$\begin{bmatrix} 1 \times 1, 256, \\ 3 \times 3, 256,\ C = 32 \\ 1 \times 1, 256 \end{bmatrix} \times 3$ | $\begin{bmatrix} 1 \times 1, 128 \\ 3 \times 3, 128,\ C = 32 \\ 1 \times 1, 128 \end{bmatrix} \times 1$ |
| conv3 | $\begin{bmatrix} 1 \times 1, 512 \\ 3 \times 3, 512,\ C = 32 \\ 1 \times 1, 512 \end{bmatrix} \times 4$ | $\begin{bmatrix} 1 \times 1, 256 \\ 3 \times 3, 256,\ C = 32 \\ 1 \times 1, 256 \end{bmatrix} \times 1$ |
| conv4 | $\begin{bmatrix} 1 \times 1, 1024 \\ 3 \times 3, 1024,\ C = 32 \\ 1 \times 1, 1024 \end{bmatrix} \times 23$ | $\begin{bmatrix} 1 \times 1, 256 \\ 3 \times 3, 256,\ C = 32 \\ 1 \times 1, 256 \end{bmatrix} \times 1$ |
| conv5 | $\begin{bmatrix} 1 \times 1, 2048 \\ 3 \times 3, 2048,\ C = 32 \\ 1 \times 1, 2048 \end{bmatrix} \times 3$ | $\begin{bmatrix} 1 \times 1, 512 \\ 3 \times 3, 512,\ C = 32 \\ 1 \times 1, 512 \end{bmatrix} \times 1$ |

**Table 2.** The backbone architectures of DeepLabv3+ [23] and auxiliary branch in the experiment of semantic segmentation task. The architecture of ResNet is from [24].

| layer name | ResNet-101 | Auxiliary branch |
|---|---|---|
| conv1 | $7 \times 7, 64, \downarrow 2$ | RB, 128, ↑ 2<br>RB, 128<br>RB, 128, ↑ 2 |
| conv2 | $3 \times 3, MP, \downarrow 2$<br>$\begin{bmatrix} 1 \times 1, 64 \\ 3 \times 3, 64 \\ 1 \times 1, 256 \end{bmatrix} \times 3$ | $\begin{bmatrix} 3 \times 3, 128 \\ 3 \times 3, 128 \end{bmatrix} \times 1$ |
| conv3 | $\begin{bmatrix} 1 \times 1, 128 \\ 3 \times 3, 128 \\ 1 \times 1, 512 \end{bmatrix} \times 4$ | $\begin{bmatrix} 3 \times 3, 256 \\ 3 \times 3, 256 \end{bmatrix} \times 1$ |
| conv4 | $\begin{bmatrix} 1 \times 1, 256 \\ 3 \times 3, 256 \\ 1 \times 1, 1024 \end{bmatrix} \times 23$ | $\begin{bmatrix} 3 \times 3, 256 \\ 3 \times 3, 256 \end{bmatrix} \times 1$ |
| conv5 | $\begin{bmatrix} 1 \times 1, 512 \\ 3 \times 3, 512 \\ 1 \times 1, 2048 \end{bmatrix} \times 3$ | $\begin{bmatrix} 3 \times 3, 512 \\ 3 \times 3, 512 \end{bmatrix} \times 1$ |

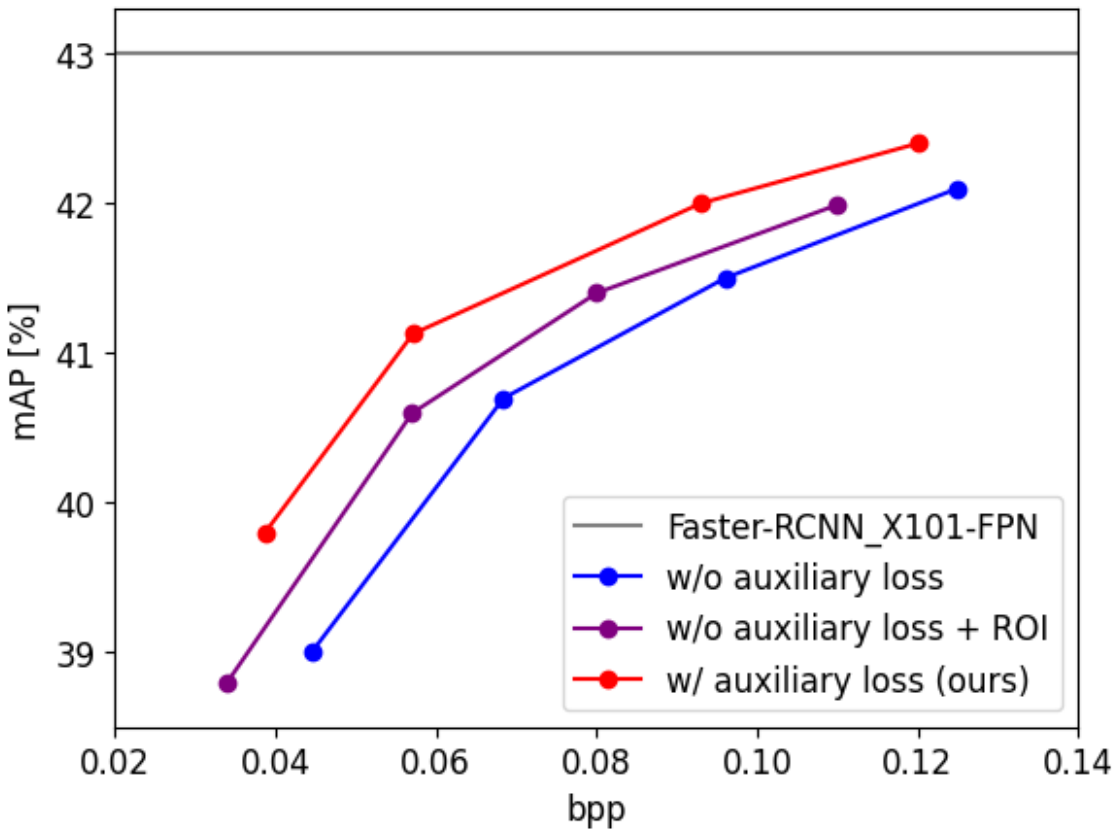

**Fig. 3.** R-D performance comparison in object detection task by Faster-RCNN.

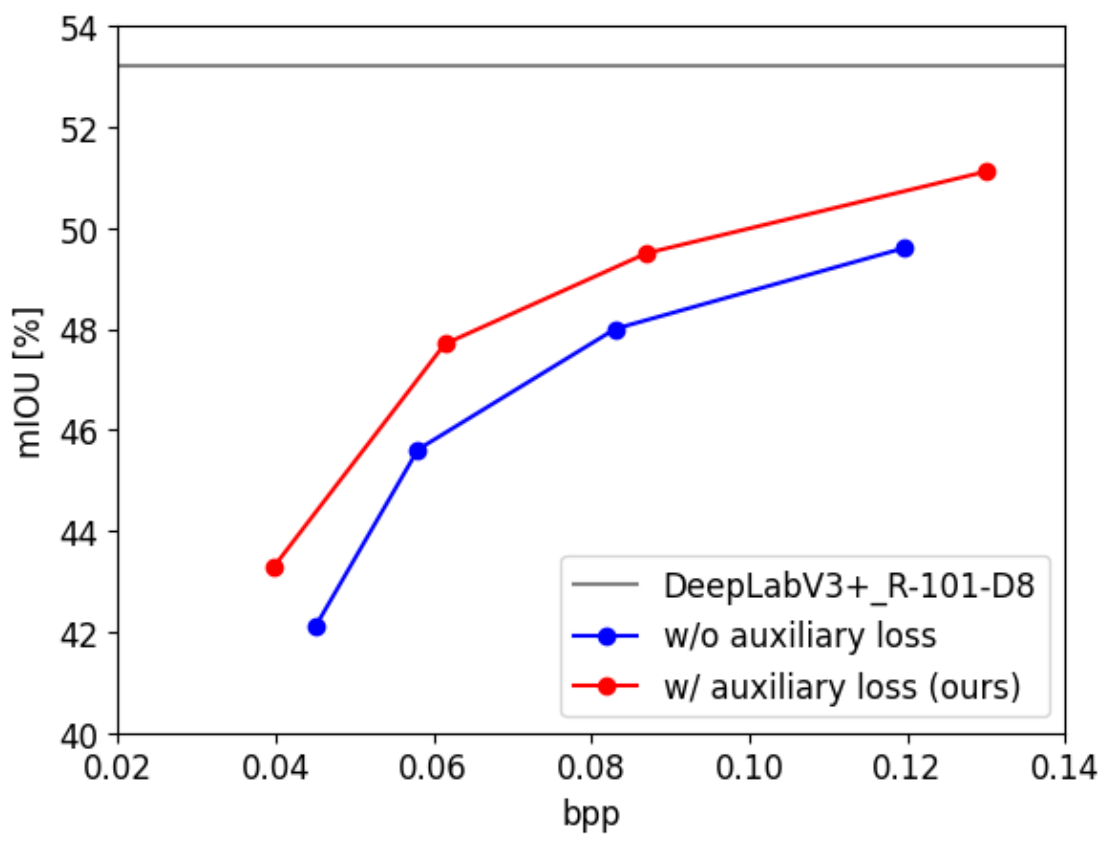

**Fig. 4.** R-D performance comparison in semantic segmentation task by DeepLabv3+.

Pascal-Context59 assigns classes to all of the image regions under evaluation, so it is not possible to define ROI; therefore, we did not use the ROI-based approach in the segmentation experiment.

**Experimental result:** The comparison results of R-D performance in object detection and semantic segmentation tasks are shown in Fig. 3 and 4. The figures demonstrate that the R-D performance improves with the addition of auxiliary loss. The Bjøntegaard Delta rate (BD-rate) improves by an average of 27.7% for object detection and by an average of 20.3% for semantic segmentation. In object detection, the proposed method shows higher R-D performance than the ROI-based approach.

To investigate the effectiveness of the proposed method, we compared the bit assignment maps, as shown in Figures 5 and 6. In the bit allocation map, brighter regions are allocated more bits. Comparing cheng2020attn [17] with w/o or w/ auxiliary loss, we can see that by training with task loss, more bits are allocated to regions that are important for the task, such as object regions or around class edges. The right side of the figures shows comparisons of the spatial tendency of bit allocation between the models of w/o and w/ auxiliary loss. The red regions indicate where the w/ aux model allocates more bits for the given image, while the blue regions show where fewer bits are allocated. We can see that the w/ aux model concentrates bits in important regions for the task and reduces the number of bits in other regions. This result shows that auxiliary loss has the effect of improving the encoder's recognition ability.

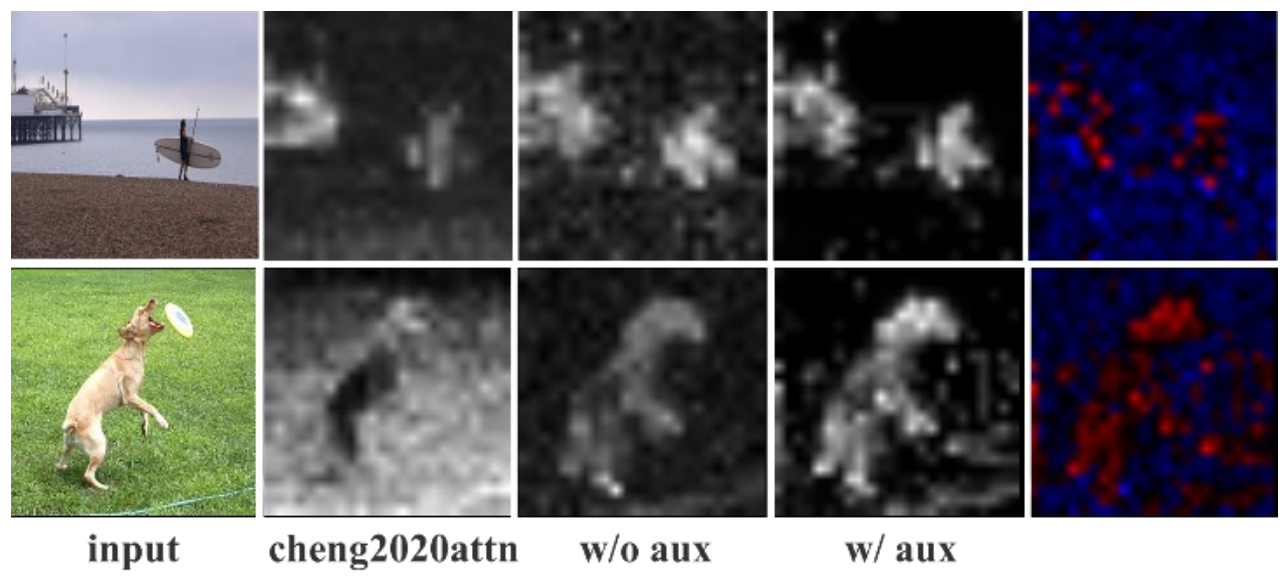

**Fig. 5.** Comparison of bit allocation. From left to right: input image from COCO2017 [25], pretrained cheng2020attn [17], w/o aux model, w/ aux model and comparision between w/o and w/ aux model.

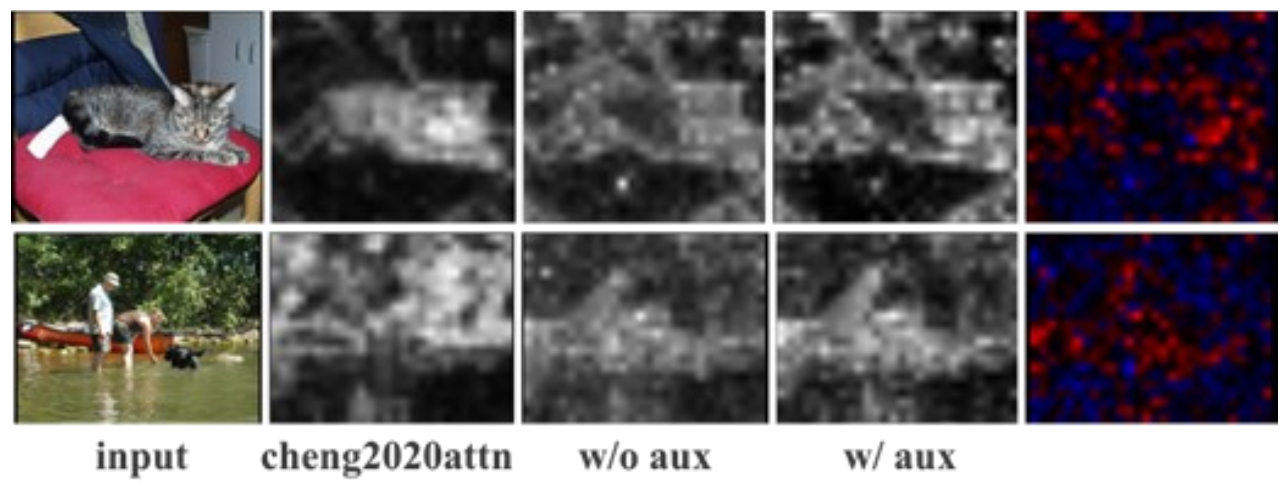

**Fig. 6.** Comparison of bit allocation. From left to right: input image from Pascal-Context59 [27], pretrained cheng2020attn [17], w/o aux model, w/ aux model and comparision between w/o and w/ aux model.

### 4.2 The position of the auxiliary branch

In the proposed method, the auxiliary branch is inserted just before the decoder. In this experiment, we changed the position of the auxiliary branch and compared the validation loss. The loss value to be compared is the main loss, equivalent to Eq. (4), which removes the auxiliary loss from Eq. (5).

**Experimental setup**: The basic experimental setup was the same as in Sec. 4.1. In this experiment, we compared validation loss with $\lambda = 0.016$ for object detection and $\lambda = 2$ for semantic segmentation. There are three types of auxiliary branch insertion positions, as shown in Fig. 6; just before the decoder (*aux-enc*: proposed), just before the final layer of the decoder (*aux-dec*), and in the middle layer of the recognition model (*aux-task*). In addition to these three types, we also compare the case without the auxiliary branch. The reason the aux-dec is placed before, rather than after, the final layer of the decoder is due to the role of the layer in the compression model. This layer reduces the dimensions to produce three RGB channels and consequently reduces the amount of information. We therefore placed it before that layer to give the effect of auxiliary loss more properly. We determined the insertion position of *aux-task* as follows. For Faster-RCNN, we inserted the auxiliary branch between conv2 and 3 of the backbone in consideration of the feature map size since there is no general configuration for setting auxiliary loss in Faster-RCNN training. For DeepLabv3+, we inserted the auxiliary branch between conv3 and conv4 of the backbone according to the MMSegmentation setting. With the change of the insertion position, we partially changed the model structure of the auxiliary branch as below. For *aux-dec*, we changed conv1 of the backbone as shown in the table. 3. For *aux-task*, we used the model after conv1 of the auxiliary branch in the table. 1 for the object detection task. In the semantic segmentation task, according to the MMSegmentation settings, we only used the head model for *aux-task*.

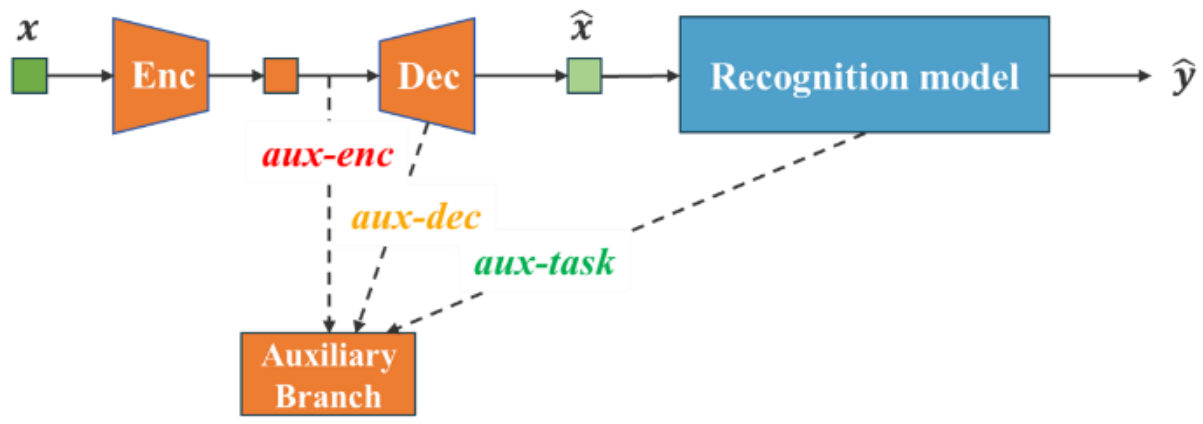

**Fig. 7.** Three types of auxiliary branch insertion positions.

**Table 3.** The difference in backbone architecture between ***aux-enc*** and ***aux-dec***. ↑ denotes upsampling.

| layer name | Auxiliary branch ***aux-enc*** | Auxiliary branch ***aux-dec*** |
|---|---|---|
| conv1 | RB, 128, ↑ 2<br>RB, 128<br>RB, 128, ↑ 2 | RB, 128, ↓ 2<br>RB, 128<br>RB, 128, |

**Experimental result**: The comparison results of validation loss in object detection and semantic segmentation tasks are shown in Fig. 7 and 8. The figures demonstrate that it is best to insert the auxiliary branch just before the decoder (*aux-enc*: proposed). It can be considered that imposing the auxiliary loss on the encoder concentratedly improves the encoder's recognition ability for the task. When inserted before the final layer of the decoder (*aux-dec*) or in the middle layer of the recognition model (*aux-task*), the auxiliary loss affects the decoder as well as the encoder. Especially in the case of *aux-dec*, the influence of auxiliary loss is stronger than that of *aux-task*, since the distance from the output layer of the auxiliary branch to the decoder is shorter. As auxiliary losses do not always help minimize main losses [28], imposing it only on the encoder could suppress the negative effect. We can also confirm that the use of auxiliary loss in both tasks has a positive effect on optimization. This provides

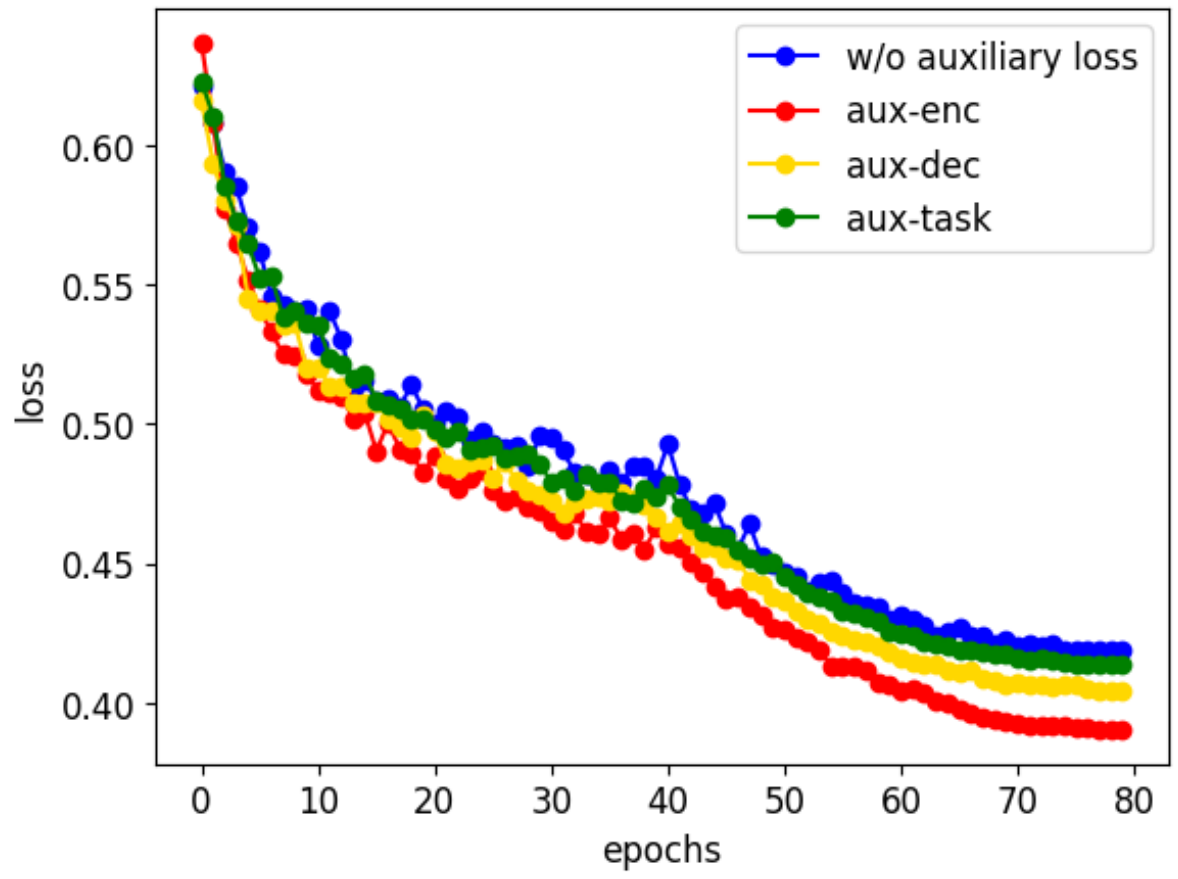

**Fig. 8.** Validation loss comparison in object detection task by Faster-RCNN.

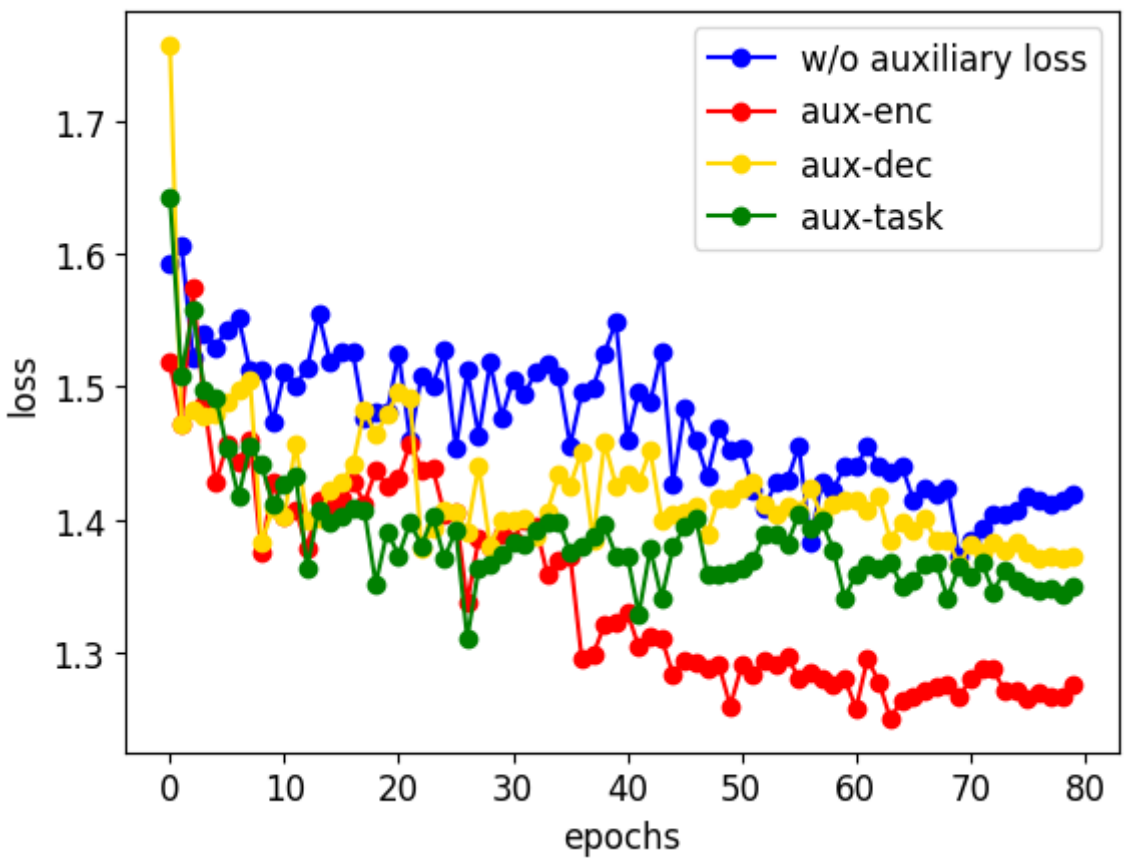

**Fig. 9.** Validation loss comparison in semantic segmentation task by DeepLabv3+.

sufficient motivation to use auxiliary losses during training of the ICM model.

## 5. CONCLUSION

In this study, we propose a novel training method for ICM models using auxiliary loss. Our proposed method imposes the auxiliary loss on the encoder of a compression model via a lightweight recognition model during training. This approach improves the encoder's recognition capability and R-D performance without any additional overhead during inference. We evaluated the proposed method by conducting experiments on object detection and semantic segmentation tasks. Compared to the conventional method without auxiliary loss, the proposed method improves the BD-rate by an average of 27.7% for object detection and 20.3% for semantic segmentation. In this study, we set the weight factor for the auxiliary loss to a fixed value for simplicity, but further performance improvement can be expected by using adaptive weight factors, such as those proposed in studies of auxiliary loss.